\documentclass[runningheads]{llncs}
\usepackage[T1]{fontenc}
\usepackage{graphicx}

\usepackage{orcidlink}
\usepackage{marvosym}

\usepackage{adjustbox}
\usepackage{amssymb}

\usepackage{graphicx}
\usepackage{amsmath}
\usepackage{comment}
\usepackage{orcidlink} 
\usepackage{graphicx}
\usepackage{multirow}
\usepackage{rotating} 
\usepackage{array} 
\usepackage{adjustbox} 

\begin{document}
\title{Generative Latent Representations of 3D Brain MRI for Multi-Task Downstream Analysis in Down Syndrome}
\titlerunning{Generative Latent Representations of 3D Brain MRI for Multi-Task Downstream Analysis in Down Syndrome}

\author{Jordi Malé\inst{1}\textsuperscript{(\Letter)}\orcidlink{0000-0003-4566-1921} \and
Juan Fortea\inst{2}\orcidlink{0000-0002-1340-638X} \and
Mateus Rozalem-Aranha\inst{2}\inst{3}\orcidlink{0000-0001-9594-292X} \and
Neus Martínez-Abadías\inst{4}\orcidlink{0000-0003-3061-2123} \and
Xavier Sevillano\inst{1}\orcidlink{0000-0002-6209-3033},
for the Alzheimer's Biomarker Consortium on Down Syndrome study}

\authorrunning{J. Male et al.}

\institute{HER - Human-Environment Research Group, La Salle - URL, Barcelona, Spain\\
\email{jordi.male@salle.url.edu}\and
Memory Unit, Department of Neurology, Institut de Recerca Sant Pau – Hospital de la Santa Creu i Sant Pau, Universitat Autònoma de Barcelona, Barcelona, Spain\\
\and
Neuroradiology Section, Department of Radiology – Hospital de la Santa Creu i Sant Pau, Universitat Autònoma de Barcelona, Barcelona, Spain\\
\and
Departament de Biologia Evolutiva, Ecologia i Ciències Ambientals (BEECA), Facultat de Biologia, Universitat de Barcelona (UB),
Barcelona, Spain\\
}

\maketitle

\begin{abstract}

Generative models have emerged as powerful tools in medical imaging, enabling tasks such as segmentation, anomaly detection, and high-quality synthetic data generation. These models typically rely on learning meaningful latent representations, which are particularly valuable given the high-dimensional nature of 3D medical images like brain magnetic resonance imaging (MRI) scans. Despite their potential, latent representations remain underexplored in terms of their structure, information content, and applicability to downstream clinical tasks. Investigating these representations is crucial for advancing the use of generative models in neuroimaging research and clinical decision-making. In this work, we develop multiple variational autoencoders (VAEs) to encode 3D brain MRI scans into compact latent space representations for generative and predictive applications. We systematically evaluate the effectiveness of the learned representations through three key analyses: (i) a quantitative and qualitative assessment of MRI reconstruction quality, (ii) a visualisation of the latent space structure using Principal Component Analysis, and (iii) downstream classification tasks on a proprietary dataset of euploid and Down syndrome individuals brain MRI scans. Our results demonstrate that the VAE successfully captures essential brain features while maintaining high reconstruction fidelity. The latent space exhibits clear clustering patterns, particularly in distinguishing individuals with Down syndrome from euploid controls. Furthermore, classification experiments on this latent space reveal the potential of generative models for encoding biologically relevant brain anatomical features. Despite the inherent trade-off between reconstruction accuracy and latent space dimensionality, our classification experiments demonstrate that generative models can effectively encode biologically relevant features of brain anatomy, offering a promising avenue for research into disorders characterised by neuroanatomical alterations.

\end{abstract}  

%
%
\section{Introduction}

Analysing medical images is crucial for accurate diagnosis and effective treatment of multiple disorders. Among the different imaging modalities, magnetic resonance imaging (MRI) stands out as a non-invasive, high-resolution, and radiation-free technique capable of capturing detailed images of diverse body tissues and organs in three dimensions (3D). In particular, brain MRI is widely used for diagnosing and monitoring a wide range of genetic, neurodevelopmental and neurodegenerative conditions.

However, human analysis and interpretation of 3D brain MRI scans are time-consuming and require extensive clinical expertise. However, with the emergence and advances in deep learning and computer vision, automation has become increasingly feasible, significantly reducing the time and effort required from specialists to tackle tasks like segmentation \cite{seg_review,Synthseg,SAM}, registration \cite{transmorph,reg_review}, and automated diagnosis \cite{Diagnosis_review,AD_Diagnosis,SZ_Diagnosis}, among others, as highlighted in recent literature surveys \cite{survey_2,survey_all}.

While deep learning—particularly supervised approaches—has achieved remarkable success in medical image analysis, its performance relies heavily on large, annotated MRI datasets. However, acquiring such datasets remains a significant challenge due to privacy concerns, ethical constraints, and the labour-intensive nature of manual annotation. Moreover, supervised models often exhibit limited generalisation beyond the scope of their training labels\cite{supervised3}.

To address these limitations, recent advancements in generative artificial intelligence have sparked significant interest in the medical imaging community. Unsupervised generative models, particularly Variational Autoencoders (VAE) and Latent Diffusion Models (LDM), have been explored for detecting brain anatomical anomalies  \cite{male_eccv_24,rumelhart_learning_1986,wolleb2022diffusion} or to generate high-quality synthetic 3D brain MRI scans \cite{pinaya2022brain,synthesis}.

A key concept in generative models is the \textit{latent space}, a lower-dimensional representation of high-dimensional data, which captures the underlying data distribution and extracts meaningful features. Such latent representation can be learned in a self-supervised or unsupervised manner by a VAE \cite{VAEs}, and exploited for multiple purposes.

In this work, we conduct a series of experiments on the latent spaces learned by a VAE from 3D brain MRI scans and analyse their reconstruction capabilities. We conduct a comprehensive analysis of the latent representations of unseen brain MRI scans and investigate how these spaces encode information relevant for performing multiple downstream tasks to support medical diagnosis. Our goal is to deepen the understanding of what these latent spaces capture and assess their capacity to represent meaningful neuroanatomical features that can serve as effective feature extraction modules for various tasks.

To achieve this goal, we utilise a proprietary dataset comprising 3D brain MRI scans from euploid (EU) subjects and individuals with Down syndrome (DS) exhibiting different levels of intellectual disability and different degrees of 
cognitive decline due to Alzheimer’s disease. To thoroughly evaluate the effectiveness of the proposed approach and to validate the relevance of the learned latent spaces, we perform a comprehensive series of experiments: \textit{(i)} a quantitative and visual evaluation of MRI reconstructions across multiple latent space dimensions to assess the fidelity and structural integrity of the generated images; \textit{(ii)} a qualitative analysis of the latent space organization through 2D projections to examine its ability to capture meaningful neuroanatomical differences among diagnostic groups; and \textit{(iii)} a quantitative assessment of multiple downstream classification tasks across varying latent space dimensionalities, demonstrating the utility of the learned representations for predicting cognitive decline associated with Alzheimer’s disease.

The main contributions of this work are threefold. First, we demonstrate the adaptability of multiple VAE architectures and their ability to generate a latent space representation that effectively supports diverse tasks, including image reconstruction, feature extraction, and classification, without requiring task-specific modifications. Second, we highlight that the learned representations are derived exclusively from brain MRI data, ensuring that all extracted features and classification signals originate purely from neuroanatomical information, free from auxiliary or confounding influences. Lastly, we show that the latent space encodes biologically meaningful differences, delivering consistently strong performance across classification tasks, with exceptional accuracy in distinguishing EU from DS subjects. These results underscore the potential of VAE-learned representations for brain MRI analysis, demonstrating their capability to capture and leverage critical neuroanatomical patterns.

\section{Diagnostic Classification based on Latent Space Representations}
\label{sec:2}

\begin{figure}[t]
\centering
\includegraphics[width=1\textwidth]{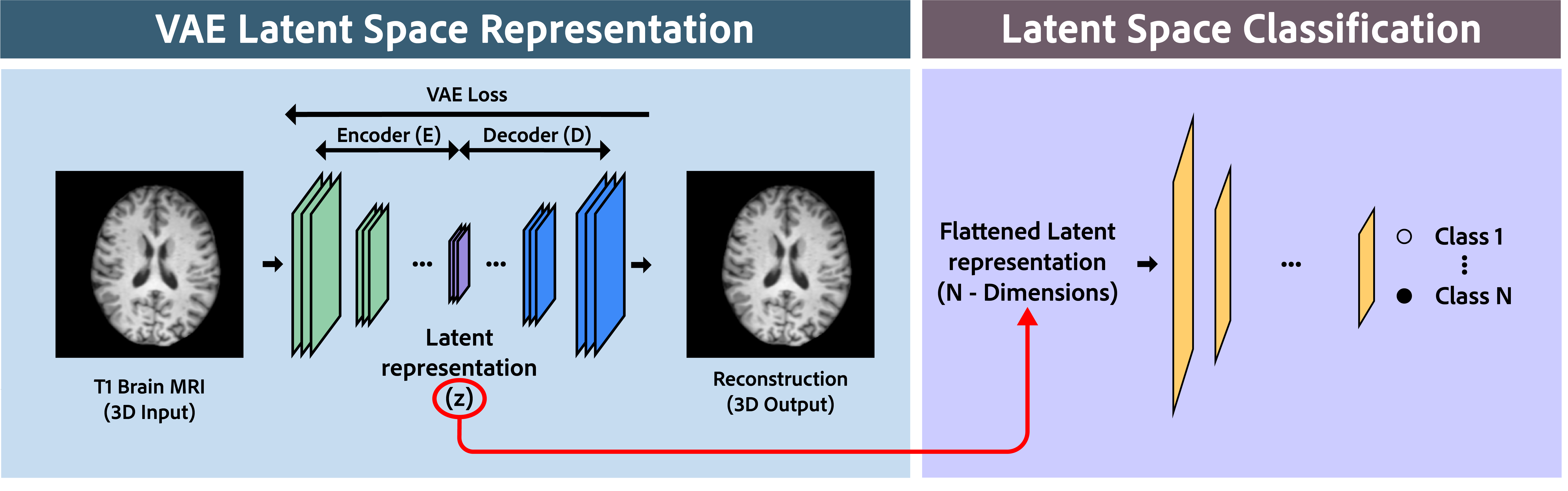}
\caption{Overview of the proposed pipeline for latent space generation and disease classification. The framework consists of two main components: A) a VAE that learns latent representations from 3D brain MRI scans, and B) a classification module that utilises these latent representations to predict a diagnosis or condition.} \label{fig:0}
\end{figure}

Our approach consists of two main stages, as depicted in Figure \ref{fig:0}. First, we train a VAE to learn meaningful latent representations of 3D brain MRI scans from multiple datasets in an unsupervised manner (Section \ref{sec:2.1}). Subsequently, these learned representations are leveraged to efficiently train downstream classification models that predict a diagnosis or condition using only the latent space representation (Section \ref{sec:2.2}). Importantly, the learned representations are not task-specific and can be exploited for a wide range of applications, including the generation of synthetic and conditionally realistic 3D brain MRI images \cite{pinaya2022brain}, anomaly detection \cite{latent_medical}, and segmentation \cite{zaman2025latentdiffusionmedicalimage}.


\subsection{Learning Latent Space Representations via Variational Autoencoders}
\label{sec:2.1}

Our compression model builds upon the work of \cite{pinaya2022brain,synthesis} and consists of a VAE trained using a combination of reconstruction loss (\( L_{\text{rec}} \)), perceptual loss (\( L_{\text{perc}} \)), and a patch-based adversarial objective (\( L_{\text{adv}} \)). In our autoencoder framework, the encoder $\mathcal{E}$ maps the input 3D brain volume \( x \in \mathbb{R}^{H \times W \times D} \) to latent distribution parameters \( (\mu, \sigma) \), from which we sample the latent representation \( z \in \mathbb{R}^{h \times w \times d \times 3} \) using the reparameterization trick \( z = \mu + \sigma \cdot \epsilon \), where \( \epsilon \sim \mathcal{N}(0, 1) \) and \( (h, w, d) \ll (H, W, D) \). The decoder $\mathcal{D}$ then reconstructs the volume as \( \hat{x} = \mathcal{D}(z) \). To regularise the latent space, we apply Kullback-Leibler divergence between the approximate posterior \( q_\mathcal{E}(z \mid x) = \mathcal{N}(z; \mu, \sigma^2) \) and a standard normal prior \( \mathcal{N}(0, 1) \), weighted by a factor $\lambda_{\text{KL}}$ to maintain reconstruction quality. The complete training objective is formulated as:

\begin{multline}
\label{eq:0}
L_{\text{AE}} = \min_{\mathcal{E}, \mathcal{D}} \max_{\psi} \Big( L_{\text{rec}}(x, \mathcal{D}(z)) + \lambda_{\text{perc}} \cdot L_{\text{perc}}(x, \mathcal{D}(z)) \\
+ \lambda_{\text{adv}} \cdot L_{\text{adv}}(\mathcal{D}(z), D_{\psi}) + \lambda_{\text{KL}} \cdot L_{\text{KL}}(\mu, \sigma) \Big)
\end{multline}

\noindent where \( L_{\text{rec}} \) is implemented as an L1 loss, \( L_{\text{perc}} \) captures high-level semantic differences using a pre-trained network, \( L_{\text{adv}} \) ensures local realism through adversarial training, and \( L_{\text{KL}} = 0.5 \sum (\mu^2 + \sigma^2 - \log(\sigma^2) - 1) \) regularises the latent space distribution. The weighting factors \( \lambda_{\text{perc}} = 2 \times 10^{-3} \), \( \lambda_{\text{adv}} = 5 \times 10^{-3} \), and \( \lambda_{\text{KL}} = 10^{-8} \) balance reconstruction quality with proper latent space regularization.

\subsection{Latent Space-based Diagnostic Classification}
\label{sec:2.2}

We hypothesise that once meaningful latent representations are learned via the VAE, they can be effectively exploited for diagnostic classification. The low dimensionality of the latent space is expected to reduce the computational complexity of the classification task while retaining essential features relevant for diagnosis. 

An additional distinctive trait of our approach is that it standardises feature extraction by leveraging the same VAE-derived latent representations for all the classification tasks, rather than training a separate feature extraction model tailored to each one of them, as performed in \cite{SZ_Diagnosis}. This ensures a more efficient and unified framework while maintaining high flexibility for different downstream classification tasks.  

As illustrated in the left part of Figure \ref{fig:0}, the classification model takes the latent representation \( z \) as input and predicts the corresponding diagnosis or condition, modeled as \( \hat{y} = f(z; \theta) \), where \( f \) represents the classifier and \( \theta \) denotes its learned parameters. All classifiers are trained using a cross-entropy loss function, which is well-suited for both binary and multi-class classification.


Given the unbalanced nature of our datasets, the classifiers loss function is designed following a unified strategy to mitigate class imbalance consisting of employing a combination of weighted cross-entropy loss and focal loss (\(\gamma=2\)), with the total loss formulated as a weighted sum: $\mathcal{L}_{\text{total}} = 0.7 \cdot \mathcal{L}_{\text{weighted}} + 0.3 \cdot \mathcal{L}_{\text{focal}}$. Weighted cross-entropy loss assigns higher importance to underrepresented classes, ensuring a balanced contribution from all categories during training. Meanwhile, focal loss \cite{focal_loss} reduces the impact of well-classified examples by down-weighting easy-to-classify instances, thereby focusing the model's learning on more challenging samples.



\section{Experimental Setup}
\label{sec:3}

\paragraph{\textbf{Datasets Description and Preprocessing.}}

To train the VAE and learn the latent space representations, we use 1,693 structural T1 MRI brain scans of euploid subjects obtained from publicly available datasets: 1.113 scans from the Human Connectome Project\footnote{Human Connectome Project, WU-Minn Consortium (Principal Investigators: David Van Essen and Kamil Ugurbil; 1U54MH091657) funded by the 16 NIH Institutes and Centers that support the NIH Blueprint for Neuroscience Research; and by the McDonnell Center for Systems Neuroscience at Washington University.} (HCP), and 580 scans from the IXI dataset \cite{IXI}. 

The classifiers used for downstream tasks are trained on entirely different datasets from those used to learn the latent spaces. In particular, each classifier is trained and evaluated using distinct subsets of a dataset provided by the Hospital Sant Pau Memory Unit (SPMU) (Barcelona, Spain), comprising a total of 931 structural T1-weighted 3D brain MRI scans encompassing both EU and DS subjects. Furthermore, 63 MRI scans from DS individuals are obtained from the public ABC-DS study dataset\footnote{Data used in this study were obtained from the Neurodegeneration in Aging Down Syndrome (NiAD) and Alzheimer’s Disease in Down Syndrome (ADDS) databases, part of the Alzheimer's Biomarkers Consortium – Down Syndrome (ABC-DS) study (\url{https://www.nia.nih.gov/research/abc-ds}, \url{niad.loni.usc.edu}. ABC-DS is a longitudinal study supported by the National Institute on Aging (NIA) and the Eunice Kennedy Shriver National Institute of Child Health and Human Development (NICHD), aimed at identifying biomarkers of Alzheimer’s disease in adults with Down syndrome. Investigators of ABC-DS contributed to study design and data collection but not to the analysis or writing of this report.} (niad.loni.usc.edu) are used to validate the trained model on external datasets from the ones seen in training.

The study is approved by the Sant Pau Hospital Research Ethics Committee (IIBSP-DOW-2014-30) and the Research Ethics Committee of Universitat Ramon Llull (CER URL 2024\_2025\_012) and is conducted in accordance with the ethical standards of the Declaration of Helsinki. Written informed consent was obtained from all participants or their legally authorised representatives before enrolment.

\begin{figure}[t]
\centering
\includegraphics[width=1\textwidth]{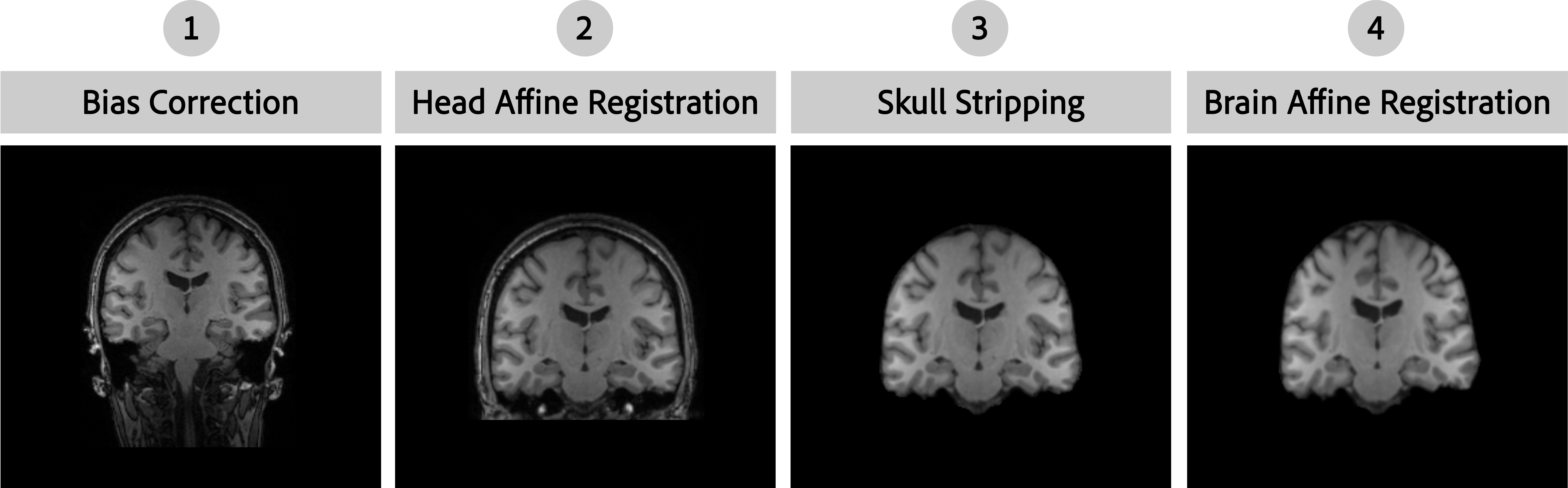}
\caption{Preprocessing pipeline for MRI scans, including bias correction, two-level affine registration and skull-stripping.}   \label{fig:1}
\end{figure}

All MRI scans undergo a preprocessing pipeline consisting of the following steps (see Figure \ref{fig:1}): \textit{i)} bias field correction, \textit{ii)} affine registration of the head to the MNI-152 template, \textit{iii)} skull-stripping using Synthstrip \cite{Synthstrip}, and \textit{iv)} affine registration of the brain to the segmented MNI-152 template. Skull-stripping is performed to segment the input brain MRI, ensuring that only brain tissue is used, excluding non-relevant anatomical structures. Additionally, all images are cropped to a standardised dimension of \( 192 \times 192 \times 192 \) voxels.  

\paragraph{\textbf{Classification Tasks.}}

For individuals with DS in the SPMU dataset, information regarding their level of intellectual disability (ID) and degree of neurodegeneration due to Alzheimer’s disease (AD) is available. Table \ref{tab:0} provides an overview of the dataset distribution across the different downstream tasks. All subgroup classifications are supervised by expert radiologists and neurologists specialising in DS and AD from Sant Pau Memory Unit.  

The first classification task involves binary classification between EU and DS subjects. For this task, we use all 931 scans from the SPMU dataset, which includes 540 EU and 391 DS subjects. Models are trained and evaluated exclusively on this dataset. To assess the generalisation capability of the trained classifiers to heterogeneous data, we conduct a second experiment. In this setting, the 63 scans from the ABC-DS dataset are used solely for testing, without retraining or fine-tuning the models.

The second classification task consists of predicting AD progression in individuals with DS, so the corresponding classifier is trained and evaluated exclusively on DS subjects (excluding EU subjects). The dataset is categorised into three subgroups based on the degree of cognitive decline due to Alzheimer’s disease:  \textit{i)} 218 DS subjects without signs of AD (no deterioration, labelled as \textit{No Det.} in Table \ref{tab:0}), \textit{ii)} 36 DS subjects with early signs of dementia (prodromal AD, labelled as \textit{Prod. AD} in Table \ref{tab:0}), and \textit{iii)} 108 DS subjects with advanced dementia symptoms (established AD, labelled as \textit{AD} in Table \ref{tab:0}). Out of the 391 total individuals with DS, 29 subjects are excluded from the analysis due to uncertain diagnoses or missing labels. Two distinct experiments are designed using these diagnostic labels. The first involves a binary classification task contrasting DS subjects without evidence of AD against those with any degree of AD-related symptoms. The second extends this framework to a multiclass classification across the three subgroups, and is aimed at assessing whether the models and their latent representations can further discriminate between stages of AD progression.

Finally, the third downstream task consists of the classification of the people with DS according to their level of ID. As in the previous task, the classifier is trained and evaluated exclusively on DS subjects, excluding EU subjects. The DS dataset is divided into three subgroups based on performance in various ID assessments: \textit{i)} 114 DS subjects with mild ID, \textit{ii)} 214 DS subjects with moderate ID, and \textit{iii)} 59 DS subjects with severe ID. Of the 391 DS subjects in the dataset, 4 are excluded from the evaluation due to missing ID evaluation results. Given the class imbalance, a binary classification experiment is conducted, contrasting subjects with mild ID against those with moderate or severe ID combined.

\begin{table}[t]
\centering
\setlength{\tabcolsep}{6pt} 
\caption{Summary of the Sant Pau Memory Unit dataset distribution across different downstream classification tasks.}
\label{tab:0}
\begin{tabular}{|c|c|c|c|c|c|c|c|}
\hline
\multirow{3}{*}{Euploid (EU)} & \multicolumn{7}{c|}{Down Syndrome (DS)} \\
\cline{2-8}
 & \multirow{2}{*}{Total} & \multicolumn{3}{c|}{AD classification} & \multicolumn{3}{c|}{ID Classification} \\
 \cline{3-8}
  &  & No Det. & Prod. AD & AD & Mild & Moderate & Severe \\
\hline
 540 & 391 & 218 & 36 & 108 & 114 & 214 & 59 \\
\hline
\end{tabular}
\end{table}

\paragraph{\textbf{Implementation Details.}} 

The VAE are implemented using the MONAI Generative framework\footnote{\url{https://github.com/Project-MONAI/GenerativeModels}} \cite{Monai}, while the classifiers are designed following a similar approach to that proposed in \cite{SZ_Diagnosis}. 

The VAE architecture follows a hierarchical structure with 3D convolutional operations. Three distinct autoencoder configurations are implemented to explore different latent space dimensions: 3$\times$3$\times$3, 12$\times$12$\times$12, and 24$\times$24$\times$24. The encoder architecture varies according to the target latent space size.

For the 24$\times$24$\times$24 latent space, the encoder consists of four stages with channel dimensions (32, 64, 128, 128) and two residual blocks per stage. For the 12$\times$12$\times$12 latent space, five stages are employed with channels (32, 64, 128, 256, 512). Finally, the 3$\times$3$\times$3 latent space requires seven stages with channels (32, 64, 128, 256, 512, 512, 1024).

Each residual block employs two 3$\times$3$\times$3 convolutions with group normalisation. Downsampling between stages is performed via strided convolutions with stride 2. The decoder mirrors this structure with corresponding upsampling stages using transpose convolutions (kernel size 3$\times$3$\times$3, stride 2), maintaining symmetrical channel dimensions. All activation functions are Swish/SiLU, except for the final output layer, which uses a sigmoid activation.

The classifiers are implemented using an architecture with residual connections, consisting of an input layer (256 neurons), three hidden layers (256, 128, 64 neurons), and an output layer with dimensions corresponding to the number of classes for each task. Each layer includes batch normalisation, ReLU activation, and dropout, with rates 0.4, 0.4, 0.3 and 0.2. Residual connections are incorporated between the first and second hidden layers, and between the second and third hidden layers. The architecture supports both binary and 3-class classification tasks.

Our code is developed using Python 3.10.14 and PyTorch, with all experiments being conducted on NVIDIA H100 GPUs on MareNostrum5 at Barcelona Supercomputing Centre (BSC).

All VAEs are trained for 1,000 epochs, using the loss objective detailed in Section \ref{sec:2.1}. The binary classifier and 3-class classifiers are trained for 50 epochs, following the training objective described in Section \ref{sec:2.2}. All models are optimised using the Adam optimiser, with a learning rate of \(5 \times 10^{-5}\) for the VAE and \(1 \times 10^{-4}\) for the classifiers.

All evaluations are conducted using 5-fold cross-validation to ensure robustness, generalizability, and mitigate overfitting, providing a more reliable estimate of real-world applicability. This results in an 80\%/20\% split for training and evaluation within each fold, which is applied independently to every experiment.

Reconstruction fidelity is evaluated using a combination of voxel-level and feature-based metrics. The Structural Similarity Index (SSIM) measures the similarity between original and reconstructed images by comparing luminance, contrast, and structural patterns, while its multi-scale extension (MS-SSIM) captures both global structure and local texture across multiple resolutions. The Mean Squared Error (MSE) quantifies the average voxel-wise intensity difference, providing a basic estimate of reconstruction error, although it does not reflect perceptual similarity. To complement these voxel-level metrics, we compute a feature distance (Feat. Dist.) based on the $\ell_2$ norm between deep embeddings extracted from a pretrained 3D MedicalNet, following the approach of \cite{pinaya2022brain}. This metric assesses whether reconstructions preserve semantically meaningful anatomical features. Finally, cosine similarity (Cos. Sim.) is employed to evaluate the angular alignment of these embeddings, with values close to one indicating strong preservation of clinically relevant structural information.

Model performance is quantitatively evaluated using accuracy, sensitivity, specificity, and the area under the receiver operating characteristic curve (AUC). For multi-class classification tasks, these metrics are computed separately for each class using one-vs-rest methodology, with macro-averaged AUC reported for overall performance assessment.

\section{Results and Discussion}

\paragraph{\textbf{VAE Reconstruction Quality}.}

The goal of this experiment is to qualitatively evaluate the brain MRI scans reconstructed by the trained VAEs with latent space representations of sizes 24$\times$24$\times$24, 12$\times$12$\times$12, and 3$\times$3$\times$3, respectively, focusing on their ability to preserve anatomical detail in the reconstructed images. 

\begin{figure}[t]
\centering
\includegraphics[width=0.8\textwidth]{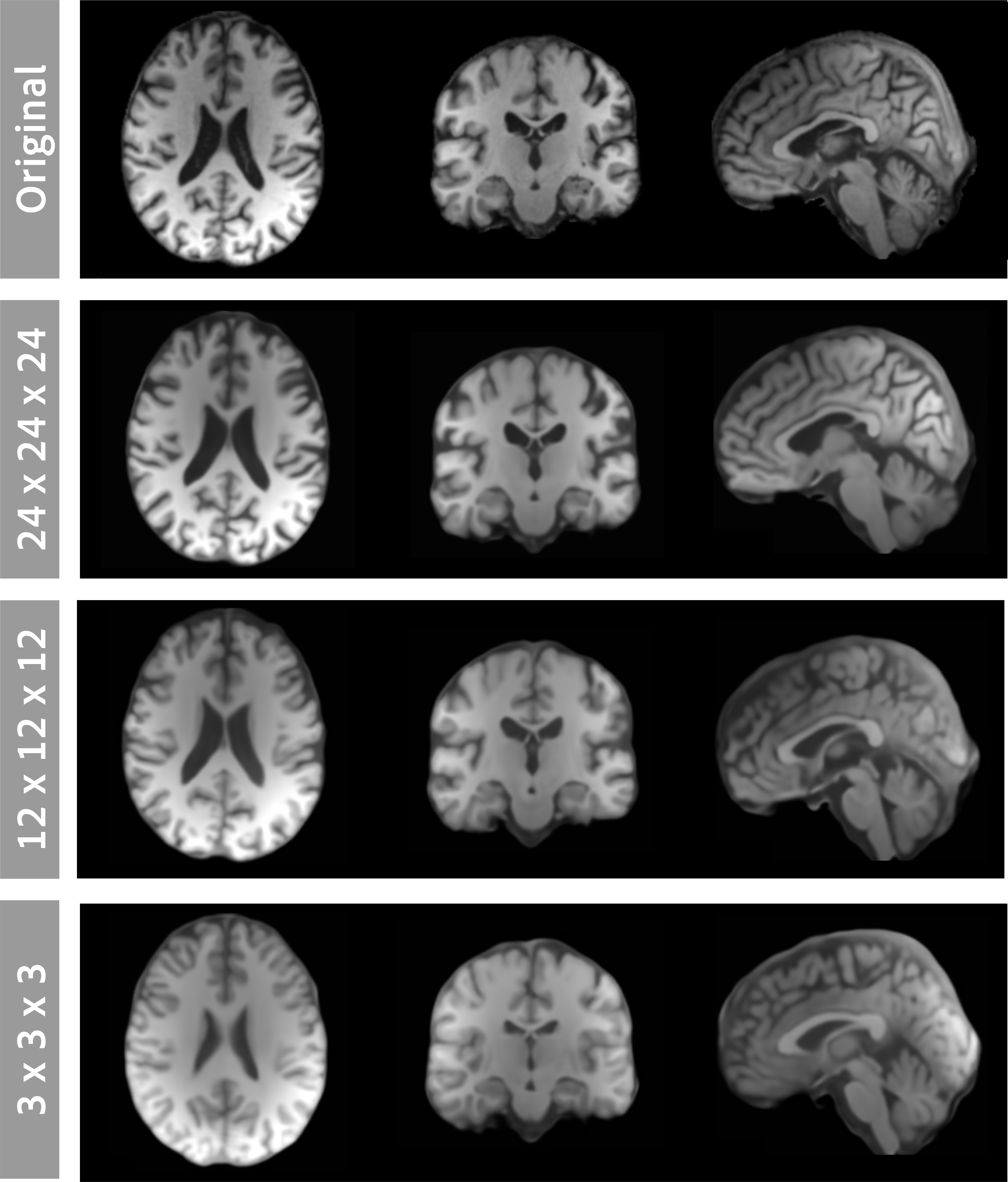}
\caption{Reconstruction of a 3D brain MRI using the trained VAE (brain reconstruction selected randomly).}
\label{fig:2}
\end{figure}

Figure \ref{fig:2} presents the reconstructed MRI outputs generated by the three VAEs. The models effectively synthesise brain structures, including cortical and subcortical regions, while maintaining anatomical symmetry, given only a latent vector as input. However, due to latent space compression and KL regularisation, fine details appear slightly smoothed, particularly in cortical boundaries and deep brain structures. The adversarial loss helps preserve high-contrast anatomical features, yet some blurring persists. While KL regularisation ensures a structured latent space that enhances generalisation for classification tasks, it also suppresses high-frequency details, limiting the model’s ability to capture fine-grained details.

\paragraph{\textbf{VAE Reconstruction Fidelity.}}  

The goal of this experiment is to quantitatively assess the extent to which latent spaces of different dimensionalities preserve anatomical detail in the reconstructed images. Table \ref{tab:rec_metrics} reports the reconstruction performance metrics of the trained VAE models, evaluated on both in-distribution (IXI) and unseen datasets (SPMU EU and SPMU DS).   

As expected after the qualitative results presented in Figure \ref{fig:2}, larger latent spaces consistently yield higher reconstruction fidelity. The 24$\times$24$\times$24 models achieve near-perfect results across all metrics, with SSIM and MS-SSIM values approaching 0.99 and negligible mean squared error, reflecting excellent structural preservation. Reducing the latent dimension to 12$\times$12$\times$12 introduces a measurable drop in performance, particularly on unseen datasets, while the most compact configuration 3$\times$3$\times$3 exhibits the largest degradation, especially in SPMU subjects. This trend indicates that the capacity of the latent space directly constrains the amount of fine-grained anatomical information retained during encoding.  

Interestingly, while IXI reconstructions remain stable across all latent dimensions (reflecting the training distribution), all SPMU subjects exhibit sharper declines as the latent space shrinks. The drop is most pronounced in DS, which is not used during training, underscoring the reduced generalisation capacity of compact latent spaces when applied to external cohorts. These results show that although all configurations preserve global anatomical structure, higher-dimensional latent spaces are more effective at maintaining fine details and generalising to unseen populations. Furthermore, these results indicate the potential for applying anomaly detection techniques based on out-of-distribution detection, as differences in reconstruction fidelity across datasets may reveal subtle distributional shifts relevant for clinical screening.

\begin{table}[t]
\centering
\setlength{\tabcolsep}{6pt}
\caption{Reconstruction fidelity performance across different latent dimensions. Metrics are reported as mean values.}
\label{tab:rec_metrics}
\begin{tabular}{|c|c|c|c|c|c|c|}
\hline
Latent Space & Dataset & SSIM $\uparrow$ & MS-SSIM $\uparrow$ & MSE $\downarrow$ & Feat. Dist. $\downarrow$ & Cos. Sim. $\uparrow$ \\
\hline
\multirow{3}{*}{24$\times$24$\times$24} 
& IXI & $.985$ & $.997$ & $.000$ & $.002$ & $1.$ \\
& SPMU EU & $.957$ & $.988$ & $.001$ & $.026$ & $.999$ \\
& SPMU DS & $.958$ & $.989$ & $.001$ & $.027$ & $.999$ \\
\hline
\multirow{3}{*}{12$\times$12$\times$12} 
& IXI & $.980$ & $.996$ & $.000$ & $.004$ & $1.$ \\
& SPMU EU & $.850$ & $.937$ & $.004$ & $.039$ & $.999$ \\
& SPMU DS & $.836$ & $.930$ & $.005$ & $.041$ & $.999$ \\
\hline
\multirow{3}{*}{3$\times$3$\times$3} 
& IXI & $.985$ & $.996$ & $.000$ & $.004$ & $1.$ \\
& SPMU EU & $.790$ & $.883$ & $.008$ & $.042$ & $.999$ \\
& SPMU DS & $.770$ & $.863$ & $.009$ & $.041$ & $.999$ \\
\hline
\end{tabular}
\end{table}

\paragraph{\textbf{Latent Space Visualization via PCA}.}

\begin{figure}[t]
\centering
\includegraphics[width=0.9\textwidth]{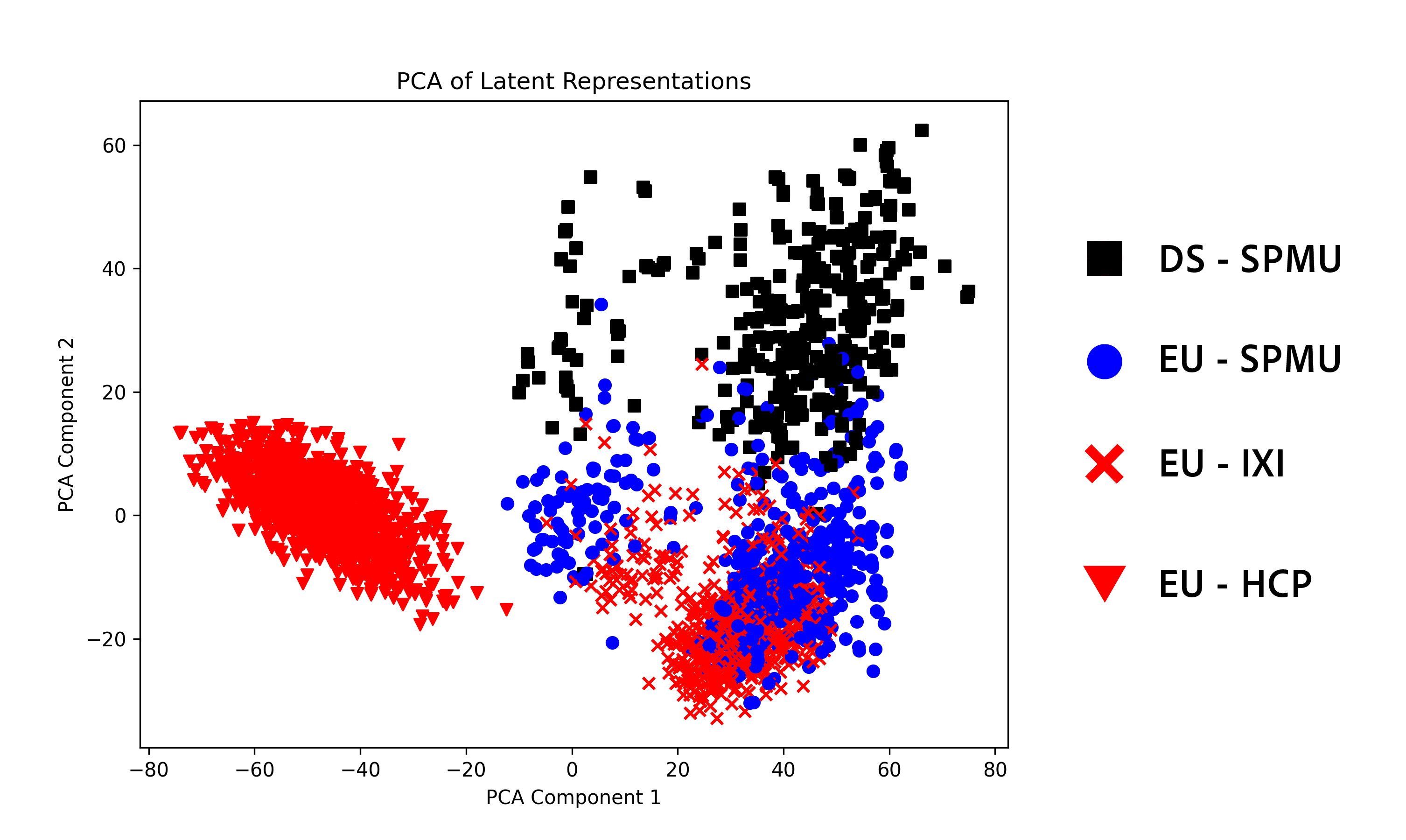}
\caption{PCA visualisation of the latent representations learned by the VAE trained on euploid subjects from the IXI dataset and the HCP dataset.}  
\label{fig:3}
\end{figure}

To assess the structure of the latent representations, we apply Principal Component Analysis (PCA) to reduce the dimensionality of the $24\times24\times24$ VAE-encoded latent space and visualise its distribution. Figure \ref{fig:3} shows the projection of the latent representations onto the first two principal components, providing insights into how well-separated different subject groups are in the learned feature space.

The scatter plot reveals distinct clusters corresponding to different subject groups. DS subjects from SPMU (black squares) form a visibly separated cluster, suggesting that the VAE effectively captures meaningful features differentiating DS from EU subjects. Among EU subjects, those from SPMU (blue circles) overlap partially with the IXI dataset (red crosses), and euploid subjects from the Human Connectome Project (HCP) dataset (red triangles) form a more compact and isolated cluster. This separation suggests that site-specific differences in acquisition protocols, demographics, or scanner characteristics contribute to variability in the learned representations. Additionally, while DS subjects exhibit a relatively compact distribution, the euploid individuals show greater overlap among themselves, indicating shared anatomical characteristics across different datasets.

This visualisation demonstrates that the VAE successfully learns meaningful latent representations that align with clinical and dataset-specific variations while reinforcing the feasibility of using these representations for downstream classification tasks.

\paragraph{\textbf{EU vs DS classification.}}  



Table \ref{tab:1} summarises the classification performance for distinguishing EU and DS subjects from latent brain MRI representations. Across all latent space configurations, the models consistently achieve near-perfect results, with mean accuracies above 97\% and AUC values exceeding 0.99. These findings demonstrate that the learned latent spaces robustly encode the discriminative features necessary for separating EU from DS subjects.  

Motivated by the latent space visualization in Figure \ref{fig:3}, which highlights the separability of the EU and DS groups, we also perform classification on the 2D space defined by the first two principal components. While performance decreases slightly (95.4\% accuracy, 0.985 AUC), the PCA space still preserves substantial discriminative power. This confirms that even after significant dimensionality reduction, the latent embeddings maintain strong separability between groups, despite information loss.

\begin{table}[t]
\centering
\setlength{\tabcolsep}{6pt} 
\caption{Mean quantitative evaluation of euploid vs. Down syndrome classification over 5-fold cross-validation.}  
\label{tab:1}
\begin{tabular}{|c|c|c|c|c|c|}
\hline
Latent Space & PCA & Accuracy & Sensitivity  & Specificity & AUC \\
\hline
24$\times$24$\times$24 & - & .994 & .998 & .987 & .998 \\
\hline
24$\times$24$\times$24 & 2 & .954 & .957 & .949 & .985 \\
\hline
12$\times$12$\times$12 & - & .996 & .998 & .992 & .999\\
\hline
3$\times$3$\times$3 & - & .973 & .981 & .962 & .995 \\
\hline
\end{tabular}
\end{table}

To further assess generalisation, we evaluate the classifiers on an external dataset comprising 63 DS subjects from the ABC-DS cohort. As shown in Table \ref{tab:abcds}, all latent spaces maintain near-perfect performance, with accuracies of 100\%, 99.71\%, and 99.00\% for the 24$\times$24$\times$24, 12$\times$12$\times$12, and 3$\times$3$\times$3 configurations, respectively. These results confirm the stability of the learned representations and their strong ability to generalise across independent cohorts, providing compelling evidence that the proposed latent embeddings capture highly reproducible and disease-relevant features. 

\begin{table}[t]
\centering
\setlength{\tabcolsep}{6pt}
\caption{Classification performance on the external ABC-DS cohort. Results reflect generalisation of the models trained on the primary dataset.}
\label{tab:abcds}
\begin{tabular}{|c|c|}
\hline
Latent Space & Accuracy \\
\hline
24$\times$24$\times$24 & 1. \\
\hline
12$\times$12$\times$12 & .997 \\
\hline
3$\times$3$\times$3 & .990 \\
\hline
\end{tabular}
\end{table}

\paragraph{\textbf{AD Stage Classification in DS Subjects.}}

Table \ref{tab:2} reports the binary classification performance for detecting Alzheimer’s disease (AD) in DS individuals. Across all latent space configurations, the models achieve consistent and robust performance, with accuracies around 83\% and AUC values close to 0.90. Importantly, all latent spaces, including the most compact 3$\times$3$\times$3 representation, preserve discriminative information sufficient to separate subjects with and without AD. Sensitivity values between 75\% and 77\% indicate reliable detection of DS subjects with AD, while specificities above 87\% highlight the models’ capacity to avoid false positives in cognitively preserved subjects. These results demonstrate that the latent spaces capture the essential structural alterations associated with AD, even under severe dimensionality reduction.  

Motivated by this strong binary performance, we extend the evaluation to a more challenging three-class setting that distinguishes between \textit{No AD}, \textit{Prodromal AD}, and \textit{Established AD}. As summarised in Table \ref{tab:3}, the models maintain reasonable overall performance (67\% to 76\% accuracy, mean AUC between 0.74 and 0.80), with high sensitivity for the No AD group ($>92\%$) and high specificity for advanced AD ($\sim$90\%). However, the Prodromal AD group remains difficult to classify reliably, reflected in low sensitivities (11\% - 20\%). This limitation likely arises from the small sample size and inherent heterogeneity of prodromal cases, which constrain the discriminative capacity of the latent space in this intermediate stage of neurodegeneration.  

These experiments indicate that the learned latent representations are consistently informative for AD detection in DS subjects. While binary classification yields robust and clinically relevant results, the extension to three classes highlights both the promise and the current limitations of latent space embeddings for fine-grained disease staging. Addressing class imbalance and incorporating complementary imaging or clinical features will be critical for improving sensitivity to early prodromal changes.

\begin{table}[t]
\centering
\setlength{\tabcolsep}{6pt} 

\caption{Quantitative evaluation of classification performance for distinguishing the presence of Alzheimer's disease in DS subjects based on latent representations of brain MRI scans.}
\label{tab:2}
\begin{tabular}{|c|c|c|c|c|}
\hline
Latent Space & Accuracy & Sensitivity & Specificity & AUC \\
\hline
24$\times$24$\times$24 & .829 & .756 & .876 & .903 \\
\hline
12$\times$12$\times$12 & .832 & .763 & .874 & .895 \\
\hline
3$\times$3$\times$3 & .834 & .770 & .871 & .895 \\
\hline
\end{tabular}
\end{table}

\begin{table}[t]
\centering
\setlength{\tabcolsep}{3pt} 

\caption{Quantitative evaluation of classification performance for 3 different stages of Alzheimer's disease progression based on latent representations of brain MRI scans.}

\label{tab:3}
\begin{tabular}{|c|c|c|c|c|c|c|c|c|}
\hline
\multirow{2}{*}{Latent Space} & \multirow{2}{*}{Accuracy} & \multicolumn{3}{c|}{Sensitivity} & \multicolumn{3}{c|}{Specificity} & \multirow{2}{*}{AUC} \\
\cline{3-8}
 & & No Det. & Prod. AD & AD & No Det. & Prod. AD & AD & \\
\hline
24$\times$24$\times$24 & $.754$ & .927 & .111 & .621 & .632 & .969 & .897 & $.802$ \\
\hline
12$\times$12$\times$12  & $.674$ & .748 & .196 & .685 & .825 & .874 & .795 & $.751$ \\
\hline
3$\times$3$\times$3  & $.765$ & .931 & .111 & .649 & .659 & .966 & .901 & $.773$ \\
\hline
\end{tabular}
\end{table}

\paragraph{\textbf{ID Classification in DS Subjects.}}

Table \ref{tab:4} reports the binary classification performance distinguishing DS subjects with mild ID from those with moderate or severe ID, based on latent representations of brain MRI scans. The best performance is observed for the 24$\times$24$\times$24 latent space, reaching an accuracy of $82.4\%$ and an AUC of $0.84$, followed closely by the 12$\times$12$\times$12 configuration.  

Across all latent space sizes, the models achieve very high sensitivity ($>94\%$ for 24$\times$24$\times$24 and 12$\times$12$\times$12), indicating that the majority of mild cases are correctly identified. However, this comes at the expense of lower specificity (ranging from $43.0\%$ to $67.5\%$), suggesting a tendency to misclassify moderate and severe subjects as mild. In contrast, the 3$\times$3$\times$3 latent space achieves a more balanced trade-off, with reduced sensitivity ($78.4\%$) but improved specificity ($67.5\%$).  

These results show that the latent space captures relevant features to discriminate mild ID from the rest, but the performance is influenced by class imbalance. The strong sensitivity highlights the model’s ability to detect mild cases, whereas the reduced specificity suggests a higher false-positive rate when distinguishing them from more impaired subjects. This evaluation provides insight into the inherent separability of mild versus moderate+severe ID in the learned latent representations.

\begin{table}[t]
\centering
\setlength{\tabcolsep}{3pt} 

\caption{Quantitative evaluation of classification performance for different levels of ID based on latent representations of brain MRI scans.}

\label{tab:4}
\begin{tabular}{|c|c|c|c|c|}
\hline
Latent Space & Accuracy & Sensitivity & Specificity & AUC \\
\hline
24$\times$24$\times$24 & .824 & .989 & .430 & .852 \\
\hline
12$\times$12$\times$12 & .811 & .949 & .483 & .827 \\
\hline
3$\times$3$\times$3 & .752 & .784 & .675 & .813 \\
\hline
\end{tabular}
\end{table}

\section{Conclusions}

This study presents a comprehensive evaluation of VAE-based latent spaces for 3D brain MRI analysis, assessing their capacity for image reconstruction, feature extraction, and downstream classification tasks in the context of Down syndrome (DS) and Alzheimer’s disease (AD).

The reconstruction experiments confirm that larger latent spaces (24×24×24) preserve fine-grained anatomical detail with near-perfect fidelity, while smaller latent spaces (12×12×12 and 3×3×3) introduce increasing levels of blurring and loss of structural information, particularly in unseen datasets. Nevertheless, all configurations are able to retain global brain structures, and the observed performance differences across datasets highlight the potential of reconstruction metrics for out-of-distribution and anomaly detection.

Latent space visualization using PCA further demonstrates that the learned embeddings capture biologically meaningful structure, with clear separability between euploid (EU) and DS subjects. These findings are quantitatively validated in the EU vs. DS classification task, where near-perfect performance is achieved across all latent spaces, with mean accuracies above 97\% and AUC values exceeding 0.99. Importantly, these results generalise robustly to an external cohort (ABC-DS), maintaining accuracies above 99\%.

Motivated by this strong performance, we evaluate disease progression in DS subjects. In the binary AD classification task (no AD vs. any AD), the models achieve accuracies around 83\% with AUC values close to 0.90, demonstrating that even highly compressed latent spaces retain discriminative features associated with AD-related neurodegeneration. Extending the task to a three-class setting (no AD, prodromal AD, established AD) shows that the models reliably identify no-AD and advanced AD cases but struggle with prodromal AD, reflecting both the intrinsic heterogeneity of early-stage disease and the small sample size available.

We further assess cognitive status by classifying DS subjects with mild intellectual disability (ID) against those with moderate or severe impairment. Here, the models demonstrate strong sensitivity (>94\% for larger latent spaces), indicating reliable detection of mild cases, but lower specificity, suggesting a tendency to overestimate impairment in more severely affected subjects. The more compact latent space (3×3×3) achieves a more balanced trade-off between sensitivity and specificity. These results indicate that the latent spaces encode relevant neuroanatomical correlates of cognitive function, though performance remains influenced by class imbalance.

Taken together, these findings demonstrate that VAE-derived latent spaces successfully capture biologically and clinically meaningful features from brain MRI scans. They provide high reconstruction fidelity, robust group separability, and consistent performance in downstream classification, generalising across datasets and diagnostic categories. While challenges remain in detecting prodromal AD and balancing ID classification due to class imbalance, these limitations point to clear avenues for future work, such as loss re-weighting, data augmentation, or multimodal integration. Moreover, the reconstruction results highlight the potential of leveraging these latent representations for anomaly detection and out-of-distribution analysis. 

Future research should explore their application to broader clinical settings, including other neurodevelopmental and neurodegenerative disorders, brain age prediction, and synthetic data generation to support data sharing and training in low-resource scenarios.

\begin{credits}
\subsubsection{\ackname}This work was partly supported by Agència de Gestió d’Ajuts Universitaris i de Recerca (AGAUR) of the Generalitat de Catalunya (2021 SGR01396, 2021 SGR00706), Agencia Española de Investigación (PID2020-113609RB-C21/AEI/ 10.13039/501100011033), the Fondation Jerome Lejeune under grant 2020b cycle-Project No.2001, and the Joan Oró grant (FI2024) from the DRU of the Generalitat de Catalunya and the European Social Fund (2024 FI-200014). Data collection and sharing for this project was supported by The Alzheimer's Disease Biomarker Consortium on Down Syndrome (ABC-DS) funded by the National Institute on Aging (NIA) and the Eunice Kennedy Shriver National Institute of Child Health and Human Development (UO1 AG051406 and UO1 AG051412). The authors thank the adults with Down syndrome volunteering as participates in this study for their invaluable contributions to this work, along with their service providers and families. We acknowledge EuroHPC Joint Undertaking for awarding the project EHPC-AI-2024A02-043 access to MareNostrum5 at Barcelona Supercomputing Center.

\subsubsection{\discintname}
The authors have no competing interests to declare that are relevant to the content of this article.
\end{credits}
%
%
%
\bibliographystyle{splncs04}
\bibliography{references}

@article{Hamner2018PediatricBD,
  title={Pediatric Brain Development in Down Syndrome: A Field in Its Infancy},
  author={Taralee Hamner and Manisha D Udhnani and Karol Osipowicz and Nancy Raitano Lee},
  journal={Journal of the International Neuropsychological Society},
  year={2018},
  volume={24},
  pages={966 - 976},
  url={https://api.semanticscholar.org/CorpusID:44074892}
}

@article{lao,
author = {Lao, Patrick and Handen, Ben and Betthauser, Tobey and Cody, Karly and Cohen, Ann and Tudorascu, Dana and Stone, Charles and Phd, Julie and Johnson, Sterling and Klunk, William and Christian, Bradley},
year = {2019},
month = {04},
pages = {1-9},
title = {Imaging neurodegeneration in Down syndrome: brain templates for amyloid burden and tissue segmentation},
volume = {13},
journal = {Brain Imaging and Behavior},
doi = {10.1007/s11682-018-9888-y}
}

@article{rodrigues,
author = {Rodrigues, Marta and Nunes, Joana and Figueiredo, Sofia and Campos, António and Geraldo, Ana},
year = {2019},
month = {12},
pages = {},
title = {Neuroimaging assessment in Down syndrome: a pictorial review},
volume = {10},
journal = {Insights into Imaging},
doi = {10.1186/s13244-019-0729-3}
}

@misc{pinaya2022brain,
      title={Brain Imaging Generation with Latent Diffusion Models}, 
      author={Walter H. L. Pinaya and Petru-Daniel Tudosiu and Jessica Dafflon and Pedro F da Costa and Virginia Fernandez and Parashkev Nachev and Sebastien Ourselin and M. Jorge Cardoso},
      year={2022},
      eprint={2209.07162},
      archivePrefix={arXiv},
      primaryClass={id='eess.IV' full_name='Image and Video Processing' is_active=True alt_name=None in_archive='eess' is_general=False description='Theory, algorithms, and architectures for the formation, capture, processing, communication, analysis, and display of images, video, and multidimensional signals in a wide variety of applications. Topics of interest include: mathematical, statistical, and perceptual image and video modeling and representation; linear and nonlinear filtering, de-blurring, enhancement, restoration, and reconstruction from degraded, low-resolution or tomographic data; lossless and lossy compression and coding; segmentation, alignment, and recognition; image rendering, visualization, and printing; computational imaging, including ultrasound, tomographic and magnetic resonance imaging; and image and video analysis, synthesis, storage, search and retrieval.'}
}

@article{latent_medical,
author = {Siddiqui, Ammar and Tirunagari, Santosh and Zia, Tehseen and Windridge, David},
year = {2025},
month = {01},
pages = {},
title = {A latent diffusion approach to visual attribution in medical imaging},
volume = {15},
journal = {Scientific Reports}}

@misc{zaman2025latentdiffusionmedicalimage,
      title={Latent Diffusion for Medical Image Segmentation: End to end learning for fast sampling and accuracy}, 
      author={Fahim Ahmed Zaman and Mathews Jacob and Amanda Chang and Kan Liu and Milan Sonka and Xiaodong Wu},
      year={2025},
      eprint={2407.12952},
      archivePrefix={arXiv},
      primaryClass={cs.CV}}

@misc{VAEs,
      title={Auto-Encoding Variational Bayes}, 
      author={Diederik P Kingma and Max Welling},
      year={2022},
      eprint={1312.6114},
      archivePrefix={arXiv},
      primaryClass={stat.ML}}

@INPROCEEDINGS{synthesis,
  author={Rombach, Robin and Blattmann, Andreas and Lorenz, Dominik and Esser, Patrick and Ommer, Björn},
  booktitle={2022 IEEE/CVF Conference on Computer Vision and Pattern Recognition (CVPR)}, 
  title={High-Resolution Image Synthesis with Latent Diffusion Models}, 
  year={2022},
  volume={},
  number={},
  pages={10674-10685}}

@misc{wolleb2022diffusion,
      title={Diffusion Models for Medical Anomaly Detection}, 
      author={Julia Wolleb and Florentin Bieder and Robin Sandkühler and Philippe C. Cattin},
      year={2022},
      eprint={2203.04306},
      archivePrefix={arXiv},
      primaryClass={id='eess.IV' full_name='Image and Video Processing' is_active=True alt_name=None in_archive='eess' is_general=False description='Theory, algorithms, and architectures for the formation, capture, processing, communication, analysis, and display of images, video, and multidimensional signals in a wide variety of applications. Topics of interest include: mathematical, statistical, and perceptual image and video modeling and representation; linear and nonlinear filtering, de-blurring, enhancement, restoration, and reconstruction from degraded, low-resolution or tomographic data; lossless and lossy compression and coding; segmentation, alignment, and recognition; image rendering, visualization, and printing; computational imaging, including ultrasound, tomographic and magnetic resonance imaging; and image and video analysis, synthesis, storage, search and retrieval.'}
}

@misc{IXI,
    author       = {Imperial College London and University College London and University of Edinburgh},
    title        = {Information eXtraction from Images (IXI) Dataset},
    howpublished = {\url{https://brain-development.org/ixi-dataset/}},
    note         = {Accessed: 2024-05-21}
}

@article{Synthstrip,
title = {{SynthStrip}: skull-stripping for any brain image},
journal = {NeuroImage},
volume = {260},
pages = {119474},
year = {2022},
issn = {1053-8119},

author = {Andrew Hoopes and Jocelyn S. Mora and Adrian V. Dalca and Bruce Fischl and Malte Hoffmann}
}

@article{supervised3,
   title={A Unifying Review of Deep and Shallow Anomaly Detection},
   volume={109},
   ISSN={1558-2256},
   number={5},
   journal={Proceedings of the IEEE},
   publisher={Institute of Electrical and Electronics Engineers (IEEE)},
   author={Ruff, Lukas and Kauffmann, Jacob R. and Vandermeulen, Robert A. and Montavon, Gregoire and Samek, Wojciech and Kloft, Marius and Dietterich, Thomas G. and Muller, Klaus-Robert},
   year={2021},
   month=may, pages={756–795} }

@misc{monai,
      title={Generative AI for Medical Imaging: extending the MONAI Framework}, 
      author={Walter H. L. Pinaya and Mark S. Graham and Eric Kerfoot and Petru-Daniel Tudosiu and Jessica Dafflon and Virginia Fernandez and Pedro Sanchez and Julia Wolleb and Pedro F. da Costa and Ashay Patel and Hyungjin Chung and Can Zhao and Wei Peng and Zelong Liu and Xueyan Mei and Oeslle Lucena and Jong Chul Ye and Sotirios A. Tsaftaris and Prerna Dogra and Andrew Feng and Marc Modat and Parashkev Nachev and Sebastien Ourselin and M. Jorge Cardoso},
      year={2023},
      eprint={2307.15208},
      archivePrefix={arXiv},
      primaryClass={id='eess.IV' full_name='Image and Video Processing' is_active=True alt_name=None in_archive='eess' is_general=False description='Theory, algorithms, and architectures for the formation, capture, processing, communication, analysis, and display of images, video, and multidimensional signals in a wide variety of applications. Topics of interest include: mathematical, statistical, and perceptual image and video modeling and representation; linear and nonlinear filtering, de-blurring, enhancement, restoration, and reconstruction from degraded, low-resolution or tomographic data; lossless and lossy compression and coding; segmentation, alignment, and recognition; image rendering, visualization, and printing; computational imaging, including ultrasound, tomographic and magnetic resonance imaging; and image and video analysis, synthesis, storage, search and retrieval.'}
}

@article{synthseg,
title = {SynthSeg: Segmentation of brain {MRI} scans of any contrast and resolution without retraining},
journal = {Medical Image Analysis},
volume = {86},
pages = {102789},
year = {2023},
issn = {1361-8415},
author = {Benjamin Billot and Douglas N. Greve and Oula Puonti and Axel Thielscher and Koen {Van Leemput} and Bruce Fischl and Adrian V. Dalca and Juan Eugenio Iglesias}
}

@article{rumelhart_learning_1986,
	title = {Learning representations by back-propagating errors},
	volume = {323},
	issn = {1476-4687},
	number = {6088},
	journal = {Nature},
	author = {Rumelhart, David E. and Hinton, Geoffrey E. and Williams, Ronald J.},
	month = oct,
	year = {1986},
	pages = {533--536},
}

@misc{seg_review,
      title={Medical Image Segmentation Review: The success of {U-Net}}, 
      author={Reza Azad and Ehsan Khodapanah Aghdam and Amelie Rauland and Yiwei Jia and Atlas Haddadi Avval and Afshin Bozorgpour and Sanaz Karimijafarbigloo and Joseph Paul Cohen and Ehsan Adeli and Dorit Merhof},
      year={2022},
      eprint={2211.14830},
      archivePrefix={arXiv},
      primaryClass={eess.IV}}

@article{SAM,
   title={Segment anything in medical images},
   volume={15},
   ISSN={2041-1723},

   number={1},
   journal={Nature Communications},
   publisher={Springer Science and Business Media LLC},
   author={Ma, Jun and He, Yuting and Li, Feifei and Han, Lin and You, Chenyu and Wang, Bo},
   year={2024},
   month=jan }

@article{transmorph,
   title={TransMorph: Transformer for unsupervised medical image registration},
   volume={82},
   ISSN={1361-8415},

   journal={Medical Image Analysis},
   publisher={Elsevier BV},
   author={Chen, Junyu and Frey, Eric C. and He, Yufan and Segars, William P. and Li, Ye and Du, Yong},
   year={2022},
   month=nov, pages={102615} }

@article{reg_review,
title = {A survey on deep learning in medical image registration: New technologies, uncertainty, evaluation metrics, and beyond},
journal = {Medical Image Analysis},
volume = {100},
pages = {103385},
year = {2025},
issn = {1361-8415},

author = {Junyu Chen and Yihao Liu and Shuwen Wei and Zhangxing Bian and Shalini Subramanian and Aaron Carass and Jerry L. Prince and Yong Du}
}

@ARTICLE{focal_loss,
  author={Lin, Tsung-Yi and Goyal, Priya and Girshick, Ross and He, Kaiming and Dollár, Piotr},
  journal={IEEE Transactions on Pattern Analysis and Machine Intelligence}, 
  title={Focal Loss for Dense Object Detection}, 
  year={2020},
  volume={42},
  number={2},
  pages={318-327},
  doi={10.1109/TPAMI.2018.2858826}}

@article{AD_Diagnosis,
author = {Al Shehri, Waleed},
year = {2022},
month = {12},
pages = {e1177},
title = {Alzheimer’s disease diagnosis and classification using deep learning techniques},
volume = {8},
journal = {PeerJ Computer Science}}

@misc{SZ_Diagnosis,
author = {Zhang, Junhao and Rao, Vishwanatha and Tian, Ye and Yang, Yanting and Acosta, Nicolas and Wan, Zihan and Lee, Pin-Yu and Zhang, Chloe and Kegeles, Lawrence and Small, Scott and Guo, Jia},
year = {2022},
month = {06},
pages = {},
title = {Detecting Schizophrenia with {3D} Structural Brain {MRI} Using Deep Learning}}

@article{Diagnosis_review,
author = {Aggarwal, Ravi and Sounderajah, Viknesh and Martin, Guy and Ting, Daniel and Karthikesalingam, Alan and King, Dominic and Ashrafian, Hutan and Darzi, Ara},
year = {2021},
month = {12},
pages = {},
title = {Diagnostic accuracy of deep learning in medical imaging: a systematic review and meta-analysis},
volume = {4},
journal = {npj Digital Medicine}
}

@article{survey_all,
title = {A survey on deep learning in medical image analysis},
journal = {Medical Image Analysis},
volume = {42},
pages = {60-88},
year = {2017},
issn = {1361-8415},
author = {Geert Litjens and Thijs Kooi and Babak Ehteshami Bejnordi and Arnaud Arindra Adiyoso Setio and Francesco Ciompi and Mohsen Ghafoorian and Jeroen A.W.M. {van der Laak} and Bram {van Ginneken} and Clara I. Sánchez}}

@article{survey_2,
author = {Celard, P. and Iglesias, E. L. and Sorribes-Fdez, J. M. and Romero, R. and Vieira, A. Seara and Borrajo, L.},
title = {A survey on deep learning applied to medical images: from simple artificial neural networks to generative models},
year = {2022},
issue_date = {Jan 2023},
publisher = {Springer-Verlag},
address = {Berlin, Heidelberg},
volume = {35},
number = {3},
issn = {0941-0643},
journal = {Neural Comput. Appl.},
month = nov,
pages = {2291–2323},
numpages = {33}
}

@misc{male_eccv_24,
      title={Towards the Discovery of Down Syndrome Brain Biomarkers Using Generative Models}, 
      author={Jordi Malé and Juan Fortea and Mateus Rozalem Aranha and Yann Heuzé and Neus Martínez-Abadías and Xavier Sevillano},
      year={2024},
      booktitle={BioImageComputing (BIC) Workshop at 2024 European Conference on Computer Vision (ECCV)}, 
      eprint={2409.13437},
      archivePrefix={arXiv},
      primaryClass={cs.CV}}

\end{document}